\def\year{2023}\relax
\documentclass[letterpaper]{article} % DO NOT CHANGE THIS
\usepackage{aaai22}  % DO NOT CHANGE THIS
\usepackage{times}  % DO NOT CHANGE THIS
\usepackage{helvet}  % DO NOT CHANGE THIS
\usepackage{courier}  % DO NOT CHANGE THIS

\usepackage[hyphens]{url}  % DO NOT CHANGE THIS
\usepackage{graphicx} % DO NOT CHANGE THIS
\usepackage{natbib}  % DO NOT CHANGE THIS AND DO NOT ADD ANY OPTIONS TO IT
\usepackage{caption} % DO NOT CHANGE THIS AND DO NOT ADD ANY OPTIONS TO IT
\DeclareCaptionStyle{ruled}{labelfont=normalfont,labelsep=colon,strut=off} % DO NOT CHANGE THIS
\usepackage{algorithm}
\usepackage{algorithmic}

\usepackage{booktabs}
\usepackage{multirow}
\usepackage{adjustbox}

\usepackage{xspace}
\usepackage{comment}
\usepackage{xcolor}
\usepackage{makecell}
\usepackage{adjustbox}
\usepackage{txfonts}
\pdfoutput=1
\usepackage{colortbl}
\usepackage{csquotes}
\usepackage{enumitem} 
\usepackage{soul}

\usepackage{newfloat}
\usepackage{listings}
\floatstyle{ruled}
\newfloat{listing}{tb}{lst}{}
\floatname{listing}{Listing}
\title{Does Human Collaboration Enhance the Accuracy of \\Identifying LLM-Generated Deepfake Texts?}

\renewcommand{\textsuperscript}[1]{\raisebox{0.5ex}{#1}}

\author{Adaku Uchendu
\thanks{Equal contribution.}$^{1,2}$,
Jooyoung Lee$^*$$^2$,
Hua Shen$^*$$^{2,4}$,
Thai Le$^3$,
\\Ting-Hao ‘Kenneth’ Huang$^2$,
Dongwon Lee$^2$ \\
\normalsize{\normalfont{  %Need to reduce this font and unbolden
$^1$ MIT Lincoln Laboratory, USA\\
$^2$ The Pennsylvania State University, USA\\
$^3$ The University of Mississippi, USA \\
$^4$ University of Michigan, USA\\}
adaku.uchendu@ll.mit.edu, \\
\{jfl5838, huashen218, txh710, dongwon\}@psu.edu, \\
thaile@olemiss.edu\\
}
}

\begin{document}

\maketitle

% === File to record the revision.
% https://docs.google.com/document/d/1kwQ0wPdoiZNekgAyHl3Arfp1jB1oHL0ZrUceHGUVwXY/edit?usp=sharing

\newcommand{\todo}[1]{{\small\textcolor{red}{ [#1 ]}}}

% Latex Color Codes: 
% http://latexcolor.com/
\definecolor{babypink}{rgb}{0.96, 0.76, 0.76}
\definecolor{bananamania}{rgb}{0.98, 0.91, 0.71}
\definecolor{babyblueeyes}{rgb}{0.63, 0.79, 0.95}

 \newcommand{\hua}[1]{{\small\textcolor{orange}{\bf [#1 --Hua]}}}
 \newcommand{\kenneth}[1]{{\small\textcolor{blue}{\bf [#1 --Ken]}}}
 \newcommand{\adaku}[1]{{\small\textcolor{purple}{\bf [#1 --Adaku]}}}

% % Choose/define the color you like! haha
% \newcommand{\adaku}[1]{{\small\textcolor{cyan}{\bf [#1 --Adaku]}}}
 \newcommand{\thai}[1]{{\small\textcolor{purple}{\bf [#1 --Thai]}}}
 \newcommand{\jy}[1]{{\small\textcolor{babyblueeyes}{\bf [#1 --Jooyoung]}}}

\newcommand{\eg}{\emph{e.g.,}\xspace}%
\newcommand{\ie}{\emph{i.e.,}\xspace}

\begin{abstract}
%\thai{pls check the document class requirement, https://www.humancomputation.com/submit.html} 
% \kenneth{The opening of the abstract can be more direct and concise. The following three sentences are a bit verbose.}
% A surge of advances in large language models (\eg GPT, T-5, LLaMA) has led to significant improvement in generating extensive, coherent sentences that resemble human writing at scale, producing so-called {\em Deepfake} texts.
% This advancement, despite its benefits, can also lead to security and privacy issues (\eg training data leakage, identity obfuscation, mis/disinformation dissemination).
% As such, it has become critically important to develop effective, practical, and scalable solutions to differentiate \emph{Deepfake} texts from human-written texts.
% 
% 
% Advances in Large Language Models (\eg GPT, LLaMA) have greatly improved the generation of coherent sentences resembling human writing on a large scale, resulting in the creation of so-called ``Deepfake'' texts. However, this progress brings potential security and privacy concerns, and it is crucial to develop efficient and scalable solutions to distinguish Deepfake texts from human-written ones.
% 
Advances in Large Language Models (e.g., GPT-4, LLaMA) have improved the generation of coherent sentences resembling human writing on a large scale, resulting in the creation of so-called {\em deepfake texts}. However, this progress poses security and privacy concerns, necessitating 
effective
%efficient and scalable -- dongwon: commented out as we don't address efficiency/scalability either
solutions for distinguishing deepfake texts from human-written ones. 
% 
% 
% 
% \kenneth{Why bother studying collaboration? Add a few sentences to motivate the problem, e.g., ``Like online fact-checking, a group can collectively detect deepfake texts in real-world situations instead of doing it individually. Although prior works studied humans' ability to detect deepfake texts, none studied how collaborations impact detection performance...''}
% 
% 
% Like online fact-checking, a group can collectively detect online misinformation instead of doing it individually.
% 
Although prior works studied humans' ability to detect deepfake texts, none has examined whether ``collaboration" among humans improves the detection of deepfake texts.
% 
% 
% 
% Toward this challenge, in this work, we investigate \emph{if human collaboration can improve performance in Deepfake text detection}.
% We test the effectiveness of collaboration in the experiment using two distinct groups: (1) non-expert individuals sourced from the AMT (Amazon Mechanical Turk) platform; and (2) people with expertise in the writing field recruited from Upwork. 
% 
% 
% 
In this study, to address this gap of understanding on deepfake texts, we conducted experiments with two groups: (1) non-expert individuals from the AMT platform and (2) writing experts from the Upwork platform.
% 
% 
% 
% \kenneth{The following claim is probably too strong, as our AMT results are less clear.}
% The results show that \textbf{collaboration can potentially improve performance of detecting Deepfake text for both non-experts (51\%) and experts (69\%) from 33\% baseline}.
% The results show that \textbf{collaboration can potentially improve performance of detecting Deepfake text for both non-experts (\ie by 6.36\%) and experts (12.76\%) compared with individual detection, respectively}.
% 
% 
%%%%
% The results show that \textbf{collaboration can potentially improve performance of detecting Deepfake text for both non-experts and experts}, by 6.36\% and 12.76\% detection accuracy compared with individual detection, respectively.
% and that the task is non-trivial as ChatGPT achieved a 38\% accuracy in the same task.\kenneth{Why mention ChatGPT here? It's probably more direct to just mention 1 baseline.}
% 
% 
% 
The results demonstrate that \textit{collaboration among humans can potentially improve the detection of deepfake texts  for both groups}, increasing detection accuracies by 6.36\% for non-experts and 12.76\% for experts, respectively, compared to individuals' detection accuracies. 
%
% % 
% We further analyzed the justifications experts and non-experts used to detect the Deepfake texts, where we find examining coherence and consistency of texts seems to be most useful in detecting deepfake texts. 
% 
% 
We further analyze the explanations that humans used for detecting a piece of text as deepfake text, and find that the strongest indicator of deepfake texts is their lack of coherence and consistency.
% 
% \kenneth{What exactly is ``submitted justification''? Use more direct/simple language.}
% Furthermore, we analyze the submitted justification distribution in regard to detection performance, where we find that examining coherence and consistency in texts is useful in detecting deepfake texts.\kenneth{I might just mention ``We further analyzed the clues workers used to detect deepfake text and found that...''}
% 
% 
% In conclusion, this study sheds light on the design of future tools/framework 
% to facilitate collaborative human detection of deepfake texts.
Our study provides useful insights for future tools and framework designs to facilitate the collaborative human detection of deepfake texts.
The experiment datasets and AMT implementations are available at: 
\url{https://github.com/huashen218/llm-deepfake-human-study.git}
% \hua{double checked the link to be correct.}

%\hua{Abstract finished, please double check, comment, and revise : )}
\end{abstract}

\section{Introduction}
\label{section:introduction}

\begin{figure}[!t]
    \centering
    \includegraphics[width=0.8\columnwidth]{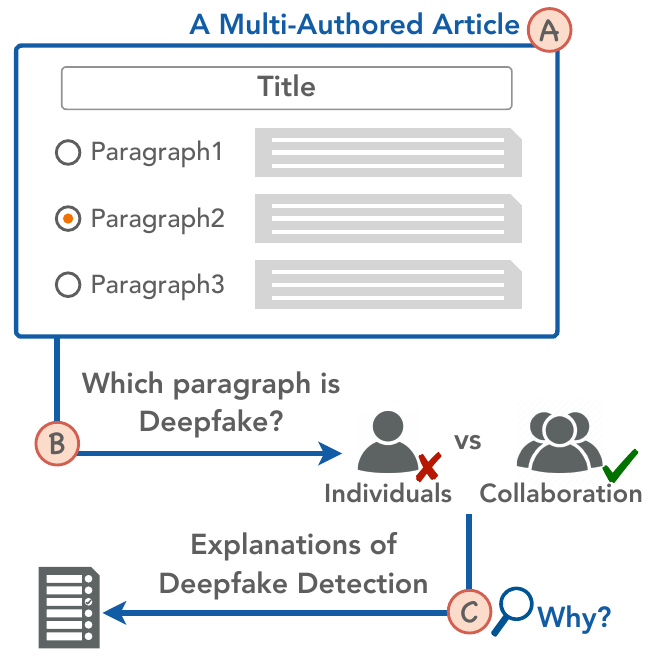}
    \caption{An overview of human studies on detecting deepfake texts. (A) A multi-authored article with 3 paragraphs, including both human-written \& LLM-generated paragraphs; (B) We conduct human studies to ask either individuals  or collaborative humans to detect  deepfake texts; (C) In-depth analysis of the categorical explanations for deepfake text detection from both groups.}
    \label{fig:task}
\end{figure}

% In recent years, AI technologies have drastically advanced, enabling the generation of human-quality artifacts in various modalities, including texts, images, and videos~\cite{fagni2021tweepfake,zhang2022deepfake,pudeepfake,shen2023parachute}.
% 
% 
% 
%\hua{Intro finished. Please feel free to proofread and revise.}
In recent years, significant advancements in AI technologies have revolutionized the generation of high-quality artifacts across various modalities, including texts, images, and videos~\cite{fagni2021tweepfake,zhang2022deepfake,pudeepfake,shen2023parachute}.
% 
% 
% These AI-generated artifacts are known as {\bf Deepfakes}.
These AI-generated artifacts, commonly referred to as Deepfakes, have garnered considerable attention.
% 
% 
% 
% In particular, the advanced {\em Natural Language Generation} (NLG) techniques with Large Language Models (LLM), such as GPT-4 \cite{OpenAI2023GPT4TR} or T5 
% \cite{raffel2020exploring}, has enabled people to generate long coherent texts without any intervention~\cite{wu2023scattershot}.
% 
% 
% 
% 
Specifically, the progress made in Natural Language Generation (NLG) techniques, leveraging Large Language Models (LLMs) like GPT-4~\cite{OpenAI2023GPT4TR} or T5~\cite{raffel2020exploring}, has facilitated the production of long and coherent machine-generated texts without human intervention~\cite{wu2023scattershot}. 
% 
% 
% In this work, we refer to such machine-generated texts as \textbf{Deepfake Texts}\footnote{This is also known as {\em neural} or {\em LLM-generated} texts.} and generative language models as {\em Neural Text Generator} (NTG), 
% as is used by \cite{zhong2020neural}.
% 
% 
% 
For the purpose of this study, we designate such neural or LLM-generated texts as \textbf{deepfake texts},
% \footnote{This is also known as {\em neural}
% or {\em LLM-generated} texts.}, 
while the generative language models themselves are referred to as Neural Text Generators (NTG)~\cite{zhong2020neural}.
% 
% 
% 
% 
%\kenneth{Cite papers that also use these terminologies, we can even say, ``..we followed the terminology coined by X et al. and call them Deepfake Texts...''}
% These NTGs have several benefits, 
%including but not limited to automated generation of blog profiles, codes in several programming languages, bulk emails or product descriptions at scale for businesses.\kenneth{We don't need to mention the upside of LLMs.}
% however, as is true for any technology, NTGs can also be misused. 
% 
% 
% 
While NTGs offer numerous benefits, it is essential to acknowledge the potential misuse associated with this technological advancement \cite{shevlane2023model}. 
% 
% 
% 
% 
% 
% % 
% For example, students may abuse NTGs in writing essay homework and get caught with plagiarism due to NTGs' memorization of training samples~\cite{lee2023language}; users can create stereotyping, misrepresenting, and demeaning content~\cite{weidinger2021ethical}; cyber attackers can generate malicious code~\cite{chen2021evaluating}; and state-backed operators may use NTGs as part of disinformation attacks~\cite{bagdasaryan2022spinning} 
% 
% 
% 
For instance, NTGs can be employed by students to complete their essay assignments, leading to potential plagiarism due to NTGs' memorization of training samples~\cite{lee2023language}. Moreover, scammers may exploit NTGs to craft sophisticated phishing messages, or stereotyping, misrepresenting, and demeaning content~\cite{weidinger2021ethical}, while malicious code generation~\cite{chen2021evaluating} and disinformation attacks by state-backed operators are also plausible scenarios~\cite{bagdasaryan2022spinning}. 
% 
%\thai{citations if available}.\kenneth{Yes, as this is one of the core motivation. Please add more cases and add citations for each of them.}
% In light of this, it is imperative to prioritize research on effective methods for distinguishing between deepfake texts and human-authored writings. 
% 
% 
Given these concerns, it becomes imperative to prioritize research efforts towards developing effective methodologies for distinguishing deepfake texts from those authored by humans.
% 
% 
% 

% 
% The field of 
% % open-ended text generation is still relatively new, 
% both computational and non-computational detection of deepfake texts have been actively studied in recent years and are well surveyed by~\citet{uchendu2022attribution}.
% 
% 
% 

Both computational and non-computational approaches for detecting deepfake texts have received significant attention in recent years~\cite{uchendu2021turingbench,clark-etal-2021-thats,dou2022gpt,brown2020language}, and have been comprehensively surveyed by~\citet{uchendu2022attribution}.
% 
% 
% 
% 
% Recent literature (e.g., \cite{uchendu2021turingbench,clark-etal-2021-thats,dou2022gpt,brown2020language}) has shown that, by and large, humans are {\em not} good at detecting deepfake texts, performing only slightly better than the level of random guessing.
% 
% 
% 
However, emerging literature~\cite{uchendu2021turingbench,dou2022gpt} suggests that humans, on average, struggle to detect deepfake texts, performing only slightly better than random guessing.
% 
% 
% 
% Even if humans are trained to detect deepfake texts, the performance has not improved significantly (e.g., \cite{clark-etal-2021-thats,dou2021scarecrow,tan2020detecting}).
% Therefore, a great need to differentiate Deepfake texts from human-written texts has naturally risen. 
Even with training, the performance of humans in deepfake text detection has shown limited improvement~\cite{clark-etal-2021-thats,dou2022gpt,tan2020detecting}. 
These findings highlight the need to explore alternative strategies, such as collaborative detection or leveraging advanced technological solutions, to address the challenges posed by deepfake texts effectively.

Online fact-checking efforts, as highlighted by~\citet{10.1145/3555143}, can be achieved collaboratively to detect online misinformation. Previous research has demonstrated that collective intelligence, often referred to as the ``wisdom of the crowd'', can surpass individual sensemaking capabilities~\cite{surowiecki2005wisdom}. Similarly, aggregating multiple human labels has also been shown to yield higher-quality results~\cite{zheng2017truth}.
However, limited attention has been given to understanding how collaboration affects the performance of deepfake detection. 
% 
% 
% 
% 
% \kenneth{If we go with the
% Therefore, the primary research goal of this study aims to \textbf{investigate the impact of human collaboration on detecting deepfake texts}.
Consequently, the primary objective of this study is to \textbf{investigate the impact of human collaboration on the detection of deepfake texts}. 
See an overview of the task presented in Figure~\ref{fig:task}, wherein we generate a three-paragraph article authored by both humans and LLM.
% 
% 
% a three-paragraph article is generated collaboratively by both humans and a Language Model (LLM). 
Individuals or collaborative human groups are then tasked with identifying the paragraph that has been generated by LLMs. Furthermore, we delve into the detailed explanations provided by humans to detect the deepfakes. It is worth noting that this deepfake detection design bears resemblance to the {\em Turing Test}.\footnote{Turing Test measures how human-like a model is. If a  model shows intelligent behavior usually attributed to a human and is thus, labeled a human, the  model is said to have passed the Turing Test.} 
As a result, our study focuses on addressing the following research questions:

\begin{itemize}
%[labelwidth=*,leftmargin=1.8em,align=left,label=\textbf{RQ\arabic*}]
    \item \textbf{RQ1:} Do collaborative teams or groups outperform individuals in deepfake text detection task? 
     \item \textbf{RQ2:} What types of reasoning explanations are useful indicators for deepfake text detection?
\end{itemize}

% To develop compelling human studies on evaluating effect of human collaboration (\ie RQ1), we study two representative stakeholder groups of online workers, including MTurk workers as English non-experts and UpWork workers as English experts (\ie we define English experts as individuals with at least a Bachelor's degree in English or related programs, see Section Methodology for the detailed filtering criteria on experts). Also, the two groups represent the conventional micro-task crowdsourcing setting and freelance marketplace setting, respectively. 
% 
% 
% 
To conduct comprehensive human studies on evaluating the effectiveness of human collaboration in deepfake text detection (\ie RQ1), we focus on two distinct stakeholder groups of online workers: 
Amazon Mechanical Turk (AMT) workers as English non-experts and Upwork workers as English experts. The term ``English experts'' refers to individuals who possess at least a Bachelor's degree in English or a related field (Please see the Methodology section for detailed filtering criteria for identifying experts). These two groups also represent the conventional micro-task crowdsourcing setting and the freelance marketplace setting, respectively.
% 
% 
% 
% 
% The further challenge next is \emph{how to enable human collaboration on the two platforms}. To empower human with collaborations for the two groups, we design an asynchronous collaboration approach for MTurk workers, and a synchronous collaboration way for Upwork workers to collaborate, respectively (See Methodology for more implementation details). 
% % 
% 
% 
The next challenge is to facilitate human collaboration on these two platforms. For AMT workers, we have devised an asynchronous collaboration approach, while for Upwork workers, a synchronous collaboration method has been implemented (please refer to the Methodology section for more information on the implementation details).
% 
% 
% 
% 
% Furthermore, we ask both groups to provide their explanations of detecting the Deepfake texts during the study (\ie RQ2), where we ask them to choose from a pre-defined set of seven explanation types or supplement their own justifications. We leverage these explanation collections to dive deeper into the reasoning behind human collaborative Deepfake detection.
% 
% 
% 
Furthermore, during the study, we request both groups to provide their explanations for detecting deepfake texts (\ie RQ2). They are given a predefined set of seven explanation types to choose from or the option to supplement their own explanations. By collecting these explanations, we aim to delve deeper into the reasoning process behind human collaborative deepfake text detection.

% By conducting the two human studies and comparing the human collaborative and individual evaluation for both Expert and Non-expert groups, we find that \textbf{human collaboration can potentially improve the performance of detecting deepfake texts for both groups.}
% % by 6.36\% among Non-experts and 12.76\% among experts
% % 
% Specifically, our summarize the key findings as follows:
% 
% 
% 
Through the execution of two human studies and a comparative analysis of human collaborative and individual evaluations within both the expert and non-expert groups, our research reveals that \textbf{human collaboration has the potential to enhance the performance of deepfake text detection for both stakeholder groups}. The key findings of our study can be summarized as follows:

\begin{itemize}
% \item Human collaboration does improve the performance in deepfake text detection by 6.36\% among non-experts and 12.76\% among experts;
\item Human collaboration leads to a 6.36\% improvement in deepfake text detection among non-experts and a 12.76\% improvement among experts;
\item The detection of deepfake texts is influenced by indicators such as ``consistency'', ``coherency'', ``common sense'', and ``self-contradiction'' issues;
% \item The strong indicators of Deepfake texts include ``consistency'', ``coherency'', ``common Sense'', and ``self-contradiction'' issues;
% \item Experts outperform non-experts in the task of deepfake detection for both individual and collaborative environments.
\item Experts outperform non-experts in both individual and collaborative scenarios when it comes to detecting deepfake texts.
\end{itemize}

Overall, this work focuses on investigating the impact of human collaboration on the detection of deepfake texts and demonstrates that collaborative efforts within representative groups yield superior results compared to individuals. The study sheds light on the underlying reasoning explanations, highlights limitations, and emphasizes the need for the development of computational and non-computational (including hybrid) tools to promote more robust and accurate detection methods.

\section{Related Work}
\label{section:literature}

\begin{figure*}[!t]
    \centering
\includegraphics[width=0.9\linewidth]{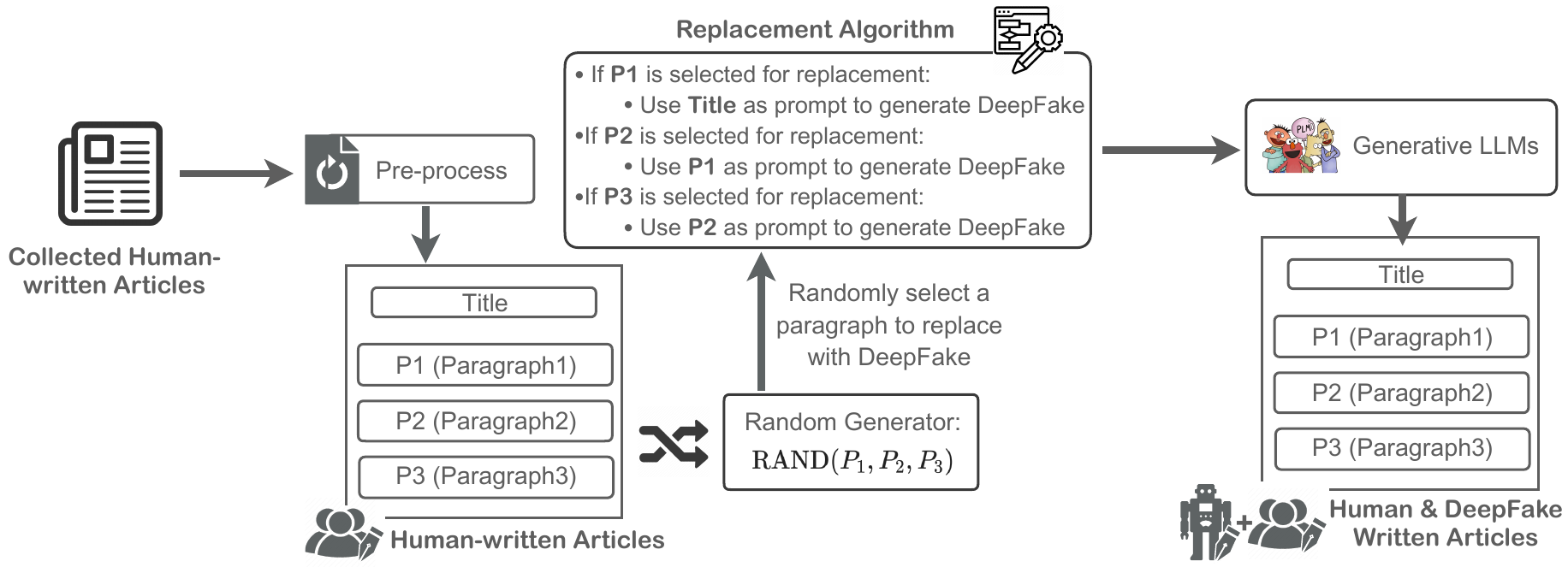}
    \caption{Illustration of the data generation process.}
    \label{fig:ntgprocess}
\end{figure*}

% \hua{I separated the 2nd literature section into two sections. Please check :) }
% \subsubsection{Human Evaluation Without Training}
\subsection{Evaluating Deepfake Texts with Laypeople}
The quality of deepfake texts has always been compared to human-written texts. Thus, since humans still remain the gold standard when evaluating machine-generated texts, 
several works have investigated human performance in 
distinguishing between human-written and machine-generated texts. 
GROVER \cite{zellers2019defending}, an NTG trained to generate news articles can easily be used maliciously. To evaluate the quality of GROVER-generated news (fake) articles, they are compared to human-written news articles. Humans are asked to 
pick which articles are more believable and GROVER-generated fake news was found to be more trustworthy \cite{zellers2019defending}.
\citet{donahue2020enabling} recruits human participants from Amazon Mechanical Turk (AMT) to detect machine-generated words in a sentence. 
\citet{uchendu2021turingbench} also recruits human participants from AMT and asks them to detect which one of two articles is machine-generated and given one article, decide if it is machine-generated or not. 
\citet{ippolito-etal-2020-automatic} evaluates the human ability to perform comparably 
given 2 different generation strategies. 
\citet{brown2020language} evaluates human performance in distinguishing human-written texts from GPT-3-generated texts. 
Finally, in all these works, the themes remain the same - humans perform poorly 
at detecting machine-generated texts, achieving about or below chance-level 
during evaluation.

% \subsubsection{Human Evaluation with Training}
\subsection{Training Humans to Evaluate Deepfake Texts}
% This has been studied in clever and nuanced ways, including training and not training. 
Since human performance in deepfake text detection is very poor, a line of studies have attempted to train the humans first and then ask them to detect the deepfake texts. For example,  \cite{gehrmann2019gltr} proposed a color-coded tool named GLTR (Giant Language Model Test Room). GLTR color codes words based on the distribution level which improves human performance from 54\% to 72\% \cite{gehrmann2019gltr}. \citet{dugan2020roft} gamifies machine-generated text detection by training humans to detect the boundary at which a document becomes deepfake to earn points. 
Humans are given the option to select one of many reasons or include their own reasons for which a sentence could be machine-generated \cite{dou2022gpt}. Our framework is modeled more closely after \citet{dugan2020roft}'s work. 
Next, \citet{clark-etal-2021-thats} proposes 3 training techniques - 
\textit{Instruction-based, Example-based,} and \textit{Comparison-based}. 
\textit{Example-based} training improved the accuracy from 50\% to 55\% \cite{clark-etal-2021-thats}. 
%Lastly, \citet{dou2022gpt} recruits human participants to annotate the error types of machine-generated texts. Participants were evaluated on an extensive qualification task which trains them \cite{dou2022gpt}. A score $\geq$ 90 out of 100 is considered a pass so the participant can move to the next round. 

Despite persistent efforts in human training, all methods except for GLTR did not yield significant improvements in human performance. However, GLTR achieved an average of 56\% F1 score on 19 pairs of human vs. state-of-the-art (SOTA) NTGs \cite{uchendu2021turingbench}, suggesting that older deepfake text detectors are inferior/obsolete to modern models. This further necessitates more thorough investigation into advanced human train methods, instead of relying on detectors. We hypothesize that previous training techniques failed because they did not consider that collaboration and skill levels could affect performance. Hence, while we implement the \textit{example-based} training technique, we also take into account expertise and collaboration elements.

%This means that while GLTR outperformed all other methods, it was built in 2019, and the results from \citet{uchendu2021turingbench} suggest that newer NTG render older deepfake text detectors inferior/obsolete. We hypothesize that previous techniques to improve human performance failed because they did not consider that collaboration and skill levels could affect performance. 
%Based on the human skill levels, they will understand hints and potentially use them differently. For instance, 
%given a task to highlight the grammar issues in a piece of text, 
%a person with college or post-college level of reading \& writing will find higher-level grammar errors (e.g., re-worded repetition, run-on sentences) than a person with 11th-grade level.
%Besides, the point of collaboration is to encourage the exchange of ideas which could de-mystify the task for humans and improve performance.
%Hence, while we implement the \textit{example-based} training technique, we also improve human performance by incorporating collaboration.

\begin{figure*}
    \centering
    \includegraphics[width=0.8 \linewidth]{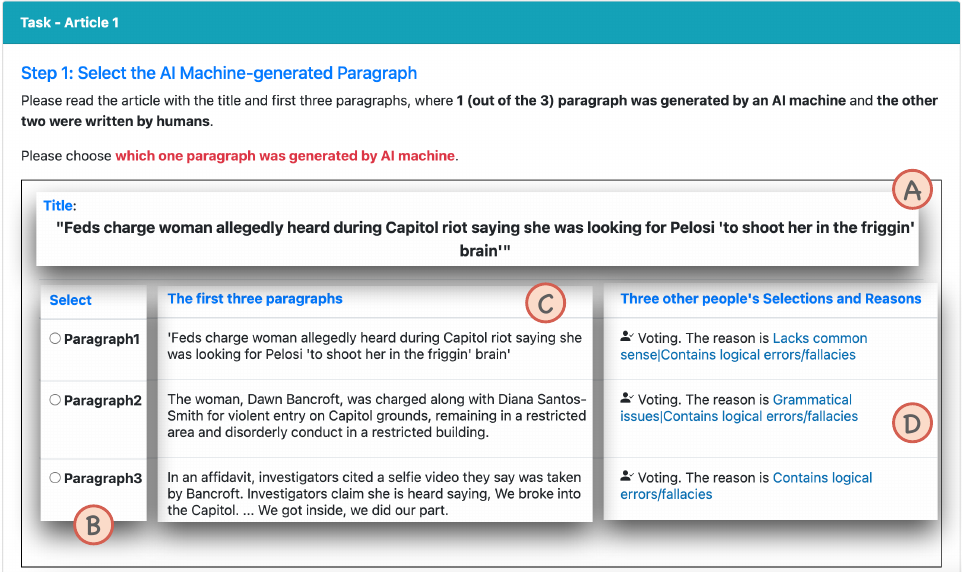}
    %\captionsetup{justification=centering} % Center the caption
    \caption{User interface for the AMT collaborative group workers to choose the LLM-generated one paragraph, whereas the individual group workers can only see A, B, and C panels.}
    \label{fig:task1}
\end{figure*}

\subsection{Automatic Evaluation of Deepfake Texts}
%\kenneth{I don't feel this is very relevant to this paper. Probably move it to the end of Related Work rather than at the beginning.}
As LLMs such as GPT-2, ChatGPT, LLaMA, etc. are able to be used
maliciously to generate misinformation at scale, 
several techniques have been 
employed to detect deepfake texts.
Using \textit{stylometric}\footnote{stylometry is the statistical analysis of an author's writing style/signature.} classifiers, researchers adopted stylometry from traditional authorship attribution solutions to achieve automatic deepfake text detection \cite{uchendu2020authorship,frohling2021feature}.
However, due to the flaws of \textit{stylometric} classifiers, \textit{deep-learning} techniques have been proposed \cite{bakhtin2019real,gpt2outputdetector,zellers2019defending,ippolito-etal-2020-automatic,ai2022whodunit,jawahar2022automatic}. 
While these \textit{deep-learning} techniques achieved high performance and significantly improved from \textit{stylometric} classifiers, they are not interpretable. To mitigate this issue, \textit{statistical-based} classifiers are proposed \cite{gehrmann2019gltr,pillutla2021information,galle2021unsupervised,pillutla2022mauve,mitchell2023detectgpt}. 
Lastly, to combine the benefits of each of the 3 types of classifiers for deepfake text detection, 
2 or more of these classifier types are combined to build a more robust classifier. \citet{uchendu2022attribution} defines these classifiers as \textit{hybrid} classifiers and they achieve superior performance \cite{liu2022coco,kushnareva2021artificial,zhong2020neural}.
Lastly, using automatic deepfake text detectors, deepfake detection has been achieved with reasonable performance. However, in the real world, as humans cannot solely depend on these models to detect deepfakes, they need to be equipped at performing the task themselves. 
A common theme in most of the detectors are that newer LLMs are harder to detect, which can sometimes make the older detectors obsolete. Thus, it is imperative that humans are also 
able to perform the task of deepfake text detection. For this reason, a few researchers have evaluated human performance in this task under several settings. See below.

% \subsection{Human Evaluation of Deepfake Texts}
% The quality of deepfake texts has always been compared to human-written texts. Thus, since humans still remain the gold standard when evaluating machine-generated texts, 
% several works have investigated human performance in 
% distinguishing between human-written and machine-generated texts. 
% This has been studied in clever and nuanced ways, including training and not training. 

\section{Methodology}
\label{sec:method}
The collective body of prior research has consistently highlighted the inherent difficulty involved in solving the deepfake detection problem~\cite{uchendu2020authorship,clark-etal-2021-thats,ippolito-etal-2020-automatic,dugan2020roft,gehrmann2019gltr}. 
Building upon the concept of ``collective intelligence" that has exhibited superior performance in online misinformation detection tasks~
\cite{horowitz2022framework,mercier2011humans, liu2018cekfakta, seo2019trust}
%\cite{horowitz2022framework,mercier2011humans, birnholtz2012tracking, liu2018cekfakta, mabrito2006study}, 
this study aims to investigate whether human collaboration can enhance the detection of deepfake texts. 
Specifically, the research methodology involves the creation of articles comprising two paragraphs authored by humans and one paragraph generated by an LLM (\ie GPT-2). 
Non-expert participants from Amazon Mechanical Turk (AMT) are then engaged in an asynchronous collaboration setting to discern the LLM-generated paragraph from the human-written paragraphs within the mixed-up articles. 
Additionally, English experts sourced from Upwork are enlisted to perform the same task but in a synchronous collaboration manner. 
To gain deeper insights into the reasoning process of humans, we analyzed the explanations provided by participants in the deepfake detection tasks. 
This study design is rooted in the practical reality that, with the increasingly impressive capabilities of LLMs, humans are increasingly inclined to employ LLMs to amend or replace portions of their own written content. 
The subsequent sections provide a detailed account of the data generation procedure, the design of the human study, and the analysis of explanations.

\begin{figure*}
    \centering
    \includegraphics[width=\linewidth]{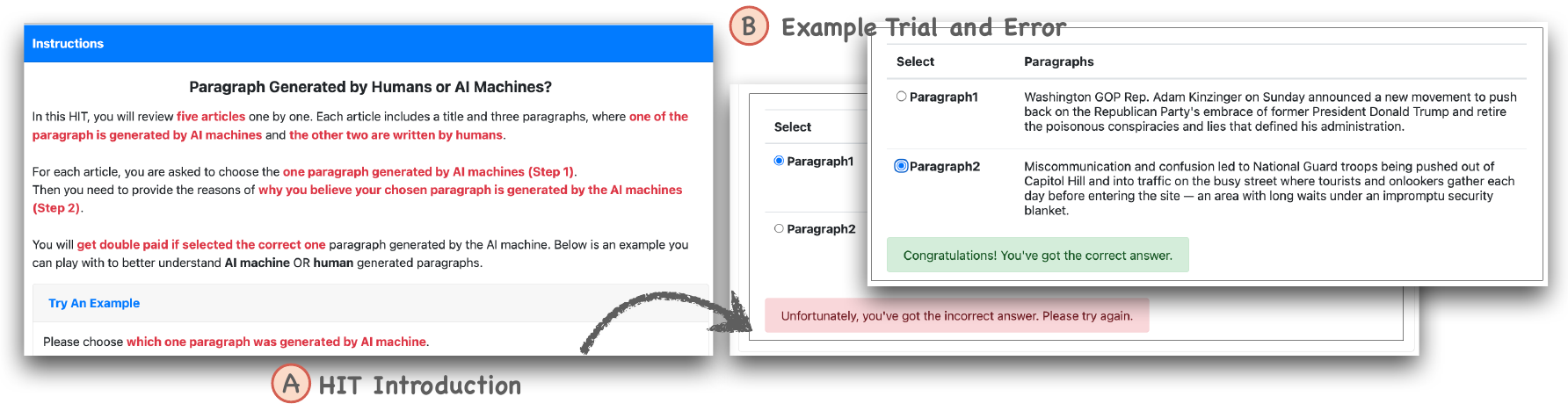}
    \caption{Instructions to train users by providing prompt feedback.}
    \label{fig:question-type2}
\end{figure*}

\subsection{Data Generation} \label{datagen}
As an overview of the data generation process shown in Figures~\ref{fig:ntgprocess}, to build this dataset, we collected 200 human-written news articles (mostly politics since this work is motivated by mitigating the risk of mis/disinformation or fake news dissemination) from reputable news sources such as CNN and Washington Post. Next, of the 200 articles, we took 
the first suitable 50 articles with at least 3 paragraphs. 
Then, we removed all paragraphs after the 3rd paragraph. 
Since the goal is to have a multi-authored article (human and LLM), we \textbf{randomly select one out of the three paragraphs} to be replaced by LLM-generated texts.
We use a random number generator to select which paragraphs are to be replaced. As a result, we replaced the \emph{Paragraph 1} in 23 articles, \emph{Paragraph 2} in 16 articles, and \emph{Paragraph 3} in 11 articles.
% 
% 

% GPT-2's \cite{radford2019language} generated texts. 
For Deepfake text generation, we used GPT-2~\cite{radford2019language} XL which has 1.5 billion parameters, and the \textit{aitextgen}\footnote{https://github.com/minimaxir/aitextgen}, a robust implementation of GPT-2 to generate texts with the default parameters\footnote{We used only GPT-2 instead of GPT-3 or above to generate the deepfake texts because: (1) GPT-2 and GPT-3 or above are using the similar algorithms. Based on~\cite{uchendu2020authorship, clark-etal-2021-thats}, human performance on detecting GPT-2 and GPT-3 texts have similar accuracies; and (2) GPT-2  is cheaper to generate texts with than GPT-3 or above since GPT-2 is open-source and GPT-3 or above is not.  }. 
We then followed the following mechanism to replace the article with the LLM-generated paragraph: 
\begin{itemize}
    \item If paragraph 1 is selected to be replaced: Use Title as a prompt to generate GPT-2 replacement;
    \item If paragraph 2 is selected to be replaced: Use Paragraph 1 as a prompt to generate GPT-2 replacement;
    \item If paragraph 3 is selected to be replaced: Use Paragraph 2 as a prompt to generate GPT-2 replacement.
\end{itemize}
Since we are unable to control the number of paragraphs GPT-2
generates given a prompt, we use a Masked Language Model (MLM) to choose the best GPT-2 replacement that fits well with the article. 
We use a BERT-base MLM \cite{devlin2018bert} to get the probability and calculate the perplexity score of the next sentence. 
Let us call this model G(.), it takes 2 inputs - the first and probable second sentence/paragraph (G($Text\_1, Text\_2$)) and outputs a score. The lower the score, the more probable $Text\_2$ is the next sentence. 
For instance, say GPT-2 texts is to replace Paragraph 2 (P2) of an article:
\begin{enumerate}
    \item We use P1 as prompt to generate P2 with GPT-2;
    \item GPT-2 generates another 3-paragraph article with P1 as the prompt;
    \item To find the suitable P2 replacement, we do G(P1, 
    each GPT-2 generated paragraph);
    \item Since low scores with G(.) is considered most probable, the P2 replacement is the GPT-2 paragraph that yielded the lowest score with G(.).
\end{enumerate}

% and got the following deepfake text replacement for 
% the paragraphs in Table\ref{tab:labels}.
After we created these multi-authored articles, we manually did a quality check of 
a few of these articles by checking for consistency and coherence. See Figure~\ref{fig:task1}(C) for an example of 
the final multi-authored article. 
We also observe that based on the replacement algorithm, some 
bias in detection may be introduced. Replacing paragraph 3 may be seen as easier 
because there is no other paragraph after it to judge the coherency. 
However, we keep the generation process fair by only using the 
text right before the paragraph as a prompt to generate the next paragraph. 
Thus, to replace paragraph 3, we only use paragraph 2 as a prompt, not 
the previous paragraphs and title.

% \begin{table}[]
% \centering
% %\footnotesize
% \begin{tabular}{r|c|c|c}
% \toprule
% \cellcolor[HTML]{DAE8FC}\textbf{Label} & \cellcolor[HTML]{DAE8FC}\textbf{Paragraph1} & \cellcolor[HTML]{DAE8FC}\textbf{Paragraph2} & \cellcolor[HTML]{DAE8FC}\textbf{Paragraph3} \\ \midrule
% \textbf{Count} & 23                  & 16                  & 11                  \\ 
% \bottomrule
% \end{tabular}
% \caption{Data labels of deepfake texts
% %\lee{need to present data in the consistent format. fonts of table 1 are larger than rest of tables}
% }
% %\vspace{-3em}
% \label{tab:labels}
% \end{table}

\subsection{Human Study Design} %\hfill

Next, as we have defined this realistic scenario, we hypothesize that collaboration will improve human detection of deepfake texts. 
Thus, we define 2 variables for this experiment -
Individual vs. Collaboration and English expert vs. English non-expert. We investigate 
how collaboration (both synchronous and asynchronous) improves from individual-based detection of deepfake texts. The hypothesis here is that when humans come together to solve 
a task, collaborative effort will be a significant improvement from average individual efforts. Additionally, as human detection of deepfake texts is non-trivial, we want to investigate if the task is non-trivial because English non-experts focus on misleading cues as opposed to English experts.

%\kenneth{I'd suggestion we just say ``Collaboration between Crowd Workers''. It's more direct.}

%\jy{Can we just say "Collaboration between Non-Experts"?, to emphasize least abt asynchronous vs. synchronous setting difference}\hua{No problem, we've revised it :)}

%\vspace{5pt}
%\noindent\textbf{Study1: Collaboration between Non-Experts.}

\subsubsection{Study1: Collaboration between AMT Participants}
%\kenneth{Why not just use subsubsection?}

%\noindent \textbf{Participants Recruitment.}

\paragraph{Participant Recruitment.}
%\kenneth{Why not just use paragraph?}
% 
% 
Inspired by \citet{clark-etal-2021-thats}, \citet{dugan2020roft}, and \citet{van2019best}, we used Amazon Mechanical Turk (AMT) to collect responses from non-expert evaluators. 
We deployed a two-stage process to conduct non-expert human studies. First, we posted a \emph{qualification-required} Human Intelligence Task (HIT) that pays $\$0.50$ per assignment on AMT to recruit 240 qualified workers.
%In terms of the qualification requirements, 
In addition to our custom qualification used for worker grouping, three built-in worker qualifications are used in all HITS, including \emph{i)} HIT Approval Rate (${\leq}98\%$), Number of Approved HITs (${\geq} 3000$), and Locale (US Only) Qualification.
Next, we only enable the qualified workers to enter the large-scale labeling tasks.
The approximate time to finish each labeling task is around 5 minutes (\ie the average time of two authors on finishing a random HIT). Therefore, we aim for $\$7.25$ per hour and set the final payment as $\$0.6$ for each assignment. Further, we provide ``double-payment'' to workers who made correct submissions as the extra bonus.

\begin{table}[!t]
\centering
% \footnotesize
\small
\begin{tabular}{c|c|c|c}
\toprule
{\textbf{Participant}} & {\textbf{Gender}} & {\textbf{Education}}         & {\textbf{Group}}              \\ \midrule \midrule
P1          & Female & Bachelor’s degree & \multirow{3}{*}{G1} \\
P2          & Female & Bachelor’s degree &                     \\
P3          & Female & Bachelor’s degree &                     \\ \cmidrule{1-4}
P4          & Female & Bachelor’s degree & \multirow{3}{*}{G2} \\
P5          & Male   & Bachelor’s degree &                     \\
P6          & Male   & Graduate degree   &                     \\ \cmidrule{1-4}
P7          & Female & Graduate degree   & \multirow{3}{*}{G3} \\
P8          & Female & Graduate degree   &                     \\
P9          & Female & Bachelor’s degree &                     \\ \cmidrule{1-4}
P10         & Female & Bachelor’s degree & \multirow{3}{*}{G4} \\
P11         & Female & Bachelor’s degree &                     \\
P12         & Male   & Bachelor’s degree &                     \\ \cmidrule{1-4}
P13         & Female & Graduate degree   & \multirow{3}{*}{G5} \\
P14         & Female & Bachelor’s degree &                     \\
P15         & Male   & Graduate degree   &                     \\ \cmidrule{1-4}
P16         & Female & Bachelor’s degree & \multirow{3}{*}{G6} \\
P17         & Male   & Bachelor’s degree &                     \\
P18         & Male   & Bachelor’s degree &                     \\ \bottomrule
\end{tabular}
    \caption{Expert (Upwork) participant demographics.}
    \label{tab:demographics}
%\vspace{-0.19in}
\end{table}

%\vspace{3pt}
%\noindent \textbf{Experiment Design.}
\paragraph{Experiment Design.}
%\kenneth{Why not just use paragraph?}
% 
During the large-scale labeling task, we divide the recruited qualified workers into two groups to represent the individual vs. collaborative settings, respectively. We define group1 as \emph{Individual Group}, in which each worker was asked to select the LLM-generated paragraph 
without any references. See Figure~\ref{fig:task1}, for example, humans in \emph{Individual Group} can only see the introduction with panels (A) (B) and (C).
On the other hand, we design group 2 to be \emph{Collaborative Group}, where the workers were asked to conduct the same task after the \emph{Individual Group} finishes all HITs (\ie see panel (A), (B), (C) in Figure~\ref{fig:task1}). In addition, workers from the \emph{Collaborative Group} could also see the selection results from group 1 in an asynchronous manner, as the example shown in Figure~\ref{fig:task1}(D), to support their own selection. 

Furthermore, we take actions to incentivize workers to provide qualified results: \emph{i)} in our instruction, we provide immediate feedback on the worker's selection to calibrate their accuracy. In specific, after reading the HIT instruction (\ie Figure~\ref{fig:question-type2} (A)), workers can get a deeper understanding of ``which paragraph is generated by AI machine'' by trial and error on selecting one example (\ie Figure~\ref{fig:question-type2} (B)). Participants were given unlimited chances to change their answers. This example-based training process was inspired by \citet{clark-etal-2021-thats}'s human evaluation study and was found to be the most effective 
training technique. \emph{ii)} We pay double compensation to the workers who provide correct answers. This aims to encourage workers to get high accuracy in selecting the correct machine-generated paragraphs. \emph{iii)} We set the minimum time constraint (\emph{i.e.,} one minute) for workers to submit their HITs so that the workers will concentrate on the task for at least one minute instead of randomly selecting one answer and submitting the HIT. Note that we also disabled the copy and paste functions in the user interface to prevent workers from searching for the paragraphs from online resources.

%\vspace{0.05in}
\subsubsection{Study2: Collaboration Between Upwork Participants}
\paragraph{Participant Recruitment.}
We utilized Upwork\footnote{Upwork is one of the leading freelance websites with a substantial network size. 
% Upwork generates 40 million monthly visits on average, and its gross services volume reached 3.5 billion dollars in 2021.
Upwork facilitates the freelance industry by introducing skilled freelancers in diverse categories like writing, design, and web development. With its automated recommendation system, we can effectively match our expert workers to our needs. See link: \url{https://sellcoursesonline.com/Upwork-statistics}.} to recruit expert evaluators, especially those with expertise in writing domains.%\thai{do you mean ``especially those with expertise in writing domains?} 
Through Upwork, we first posted a task description as a client to gather participants. We mentioned in the description that this is for research and provided all necessary information such as research objectives and example questions. Our recruitment advertisement also highlighted the mandatory requirements: (1) a participant should be at least 18 years old; and (2) a participant should be a native English speaker. Lastly, if they were willing to proceed, they were asked to submit a proposal answering the following questions: 
(1) What is the highest level of degree you have completed in school?; 
(2) Did you major in English or English Literature?; and 
(3) Describe your recent experience with similar projects.  

One useful feature for accelerating the recruitment process in Upwork is that not only workers %\thai{should we use ``workers" instead? ``employees" can be controversal.} 
can apply to the postings but also clients like us can invite prospective candidates that seem suitable for the task to submit proposals. We manually reviewed workers’ profile descriptions who specified their skill sets as copywriting, editing/proofreading, and content writing and then sent them invites. 

While making recruitment decisions, we verified participants' eligibility by checking their self-reported age, language, and education in the profile, in addition to evaluating their proposal responses. It resulted in a total of 18 finalists to officially begin the study. Next, we sent them the consent form via the platform's messaging function and activated Upwork contracts only after they returned the signed form. A primary purpose of the contracts was for clients to compensate workers based on submitted hours through the Upwork system. Participants' requested hourly wages ranged from \$25-\$35 per hour depending on their prior experiences and education levels. All 18 individuals successfully signed both documents and were compensated accordingly. Table \ref{tab:demographics} gives the self-reported demographic of recruited Upworkers.

\begin{figure}[!t]
    \centering
    \includegraphics[width=\linewidth]{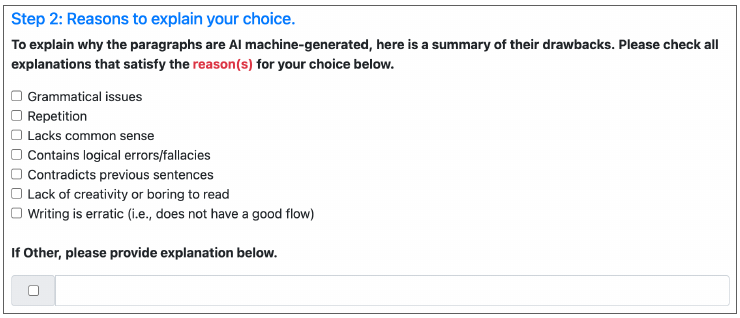}
    %\captionsetup{justification=centering} % Center the caption
    \caption{User interface for participants to select explanations for identified deepfake paragraphs.}
    \label{fig:question-type1}
\end{figure}

% \vspace{3pt}
% \subsubsection{Upwork} 
\paragraph{Experiment Design.}
To compare experts' deepfake text detection accuracy with respect to individual vs. collaborative settings, our Upwork study consists of two sub-experiments. 
The first experiment asks Upwork participants to perform a given task on their own. The second experiment requires three individuals to solve the questions as one group in a synchronous manner. We used Qualtrics\footnote{\url{https://www.qualtrics.com}} service to generate and disseminate the study form. Upwork participants were given one week to complete the survey. Upon completion, we randomly grouped 3 participants per team, resulting in 6 teams in total for synchronous collaboration (Table \ref{tab:demographics}). All discussions were conducted on the video communications software - Zoom and we leveraged Zoom's built-in audio transcription feature, which is powered by Otter.ai\footnote{\url{https://otter.ai}} for discourse analyses. In addition to the written consent obtained during the recruitment procedure, verbal consent for participation in the discussion and for audio recording was obtained prior to the start of each session. One member of the study team served as a moderator for the meetings. Depending on the participant's schedule and level of commitment in their group, each meeting lasted 1.5 - 3 hours.

\begin{table}[!t]
\centering
\small
    \centering
    \begin{adjustbox}{max width=\columnwidth}
    \begin{tabular}{c|c|c}
    \toprule
    {\textbf{Setting}}     &    {\textbf{Mean Accuracy}}                &  {\textbf{P-Value}}  \\                                                
                                  \midrule \midrule
    Baseline vs. Individual      & \multicolumn{1}{c|}{33.33\% vs. 44.99\%***}    &  3.8e-05 \\ \cmidrule{1-3}
    Baseline vs. Collaboration   & \multicolumn{1}{c|}{33.33\% vs. \textbf{51.35\%}***}    &  2.8e-05 \\ \cmidrule{1-3}
    Individual vs. Collaboration & \multicolumn{1}{c|}{44.99\% vs. \textbf{51.35\%}} &  0.054    \\ \bottomrule
    \end{tabular}
    \end{adjustbox}
    %\captionsetup{justification=centering} % Center the caption
    \caption{Paired t-test results for AMT experiments. \\ 
     (***: \textit{p} $<$ 0.001, **: \textit{p} $<$ 0.01, *: \textit{p} $<$ 0.05)}
    \label{tab:rq1_amt}
\end{table}

% \subsection{Evaluation and Analysis Method} %\hfill
\subsection{In-depth Analysis on Detection Explanations} %\hfill

We build the explanation section similar to  RoFT \cite{dugan2020roft}, a gamification technique for improving human performance in deepfake text detection.
In the RoFT framework, participants were asked to select from a pre-defined list one or more reasons such as repetition, grammar errors, etc. Participants were also
given another option, where they can enter their own justification if they do not find any suitable selection from the provided list.
%\thai{do we need to use the word ``gamification" here since this component does not have any rewards/competition?}

% \vspace{5pt}

To determine the list of pre-defined reasoning explanations in deepfake text detection, we first refer to \citet{dou2022gpt}, which provides a detailed list of 10 errors in which 
annotators have been indicated to be good indicators of deepfake texts.
However, these errors are general errors and thus some are not applicable to the task of 
detecting deepfake paragraphs. 
Therefore, due to this novel application, we 
only select the most relevant errors. Additionally, we also include relevant errors 
from \citet{dugan2020roft} including the selection of other, a gamification of deepfake text detection. 
% \vspace{5pt}
% \noindent\textbf{In-depth Evaluation on Detection Explanations.}
As the result shown in Figure~\ref{fig:question-type1}, we consequently provide seven pre-defined rationales that correspond to flaws typically observed in deepfake texts \cite{dou2022gpt,dugan2020roft}, including ``Grammatical issues'', ``Repetition'', ``Lacks common sense'', ``Contains logical errors'', ``Contradicts previous sentences'', ``Lack of creativity or boring to read'', ``Writing is erratic'' (i.e. does not have a good flow), and an additional open-ended selection - other for participants to write more of their own.
% 
% 
% These errors are general errors and thus some are not applicable to the task of 
% detecting deepfake paragraphs. 
% 
% Collecting all these error types, resulted in the 7 pred-defined error types, and 
% other. 
% 
% \hua{[TODO]Hiii Adaku, could you please help add ``how we get these justifications'' process? Thanks!}
% 

Given the pre-defined reasoning explanation list, we ask both individuals and collaborative groups to provide their explanations for each corresponding detection instance. We apply this implementation for both non-expert and expert groups, resulting in the in-depth explanation analysis with respect to four scenarios (\ie individual-expert, collaboration-expert, individual-non-expert, collaboration-non-expert). To provide more fine-grained insights, we further separate the deepfake detection results into correct detection and incorrect detection subgroups.

\section{Experimental Results}
\label{section:quant_anal}
% \hua{We can probably frame our ``Result'' Section into two parts: 
% % 
% - 1) Experiment1: Non-Expert Human Study (Including individual and collab performance as one subsection; justification patterns as another subsection ); 
% % 
% 2) Experiment2: Expert Human Study. (Including individual and collab performance as one subsection; justification patterns as another subsection ); }

\begin{table}[!t]
\small
    \centering
    \begin{adjustbox}{max width=\columnwidth}
    \begin{tabular}{c|c|c}
    \toprule
    {\textbf{Setting}}     &    {\textbf{Mean Accuracy}}                &  {\textbf{P-Value}}  \\                                                
                                  \midrule \midrule
    Baseline vs. Individual      & \multicolumn{1}{c|}{33.33\% vs. 56.11\%***}    &  8.2e-11 \\ \cmidrule{1-3}
    Baseline vs. Collaboration   & \multicolumn{1}{c|}{33.33\% vs. \textbf{68.87\%}***}    &  1.2e-12 \\ \cmidrule{1-3}
    Individual vs. Collaboration & \multicolumn{1}{c|}{56.11\% vs. \textbf{68.87\%}***} &  1.3e-05    \\ \bottomrule
    \end{tabular}
    \end{adjustbox}
    %\captionsetup{justification=centering} % Center the caption
    \caption{Paired t-test results for Upwork experiments. \\ 
     (***: \textit{p} $<$ 0.001, **: \textit{p} $<$ 0.01, *: \textit{p} $<$ 0.05)}
    \label{tab:rq1_Upwork}
\end{table}

\subsection{Evaluation Metrics and Baselines}

\paragraph{Objective Metrics.}
We measure how well participants perform the tasks and compared them across different experiment settings. To quantify the detection performance of each setting, we computed the proportion of people who got the answer correct given a set of 50 questions $Q$=\{$q_1$, $q_2$,..., $q_{50}$\}. Suppose $l_n$ is the number of participants with correct answers, and $m_n$ is the total number of participants for the question $q_n$, we calculated the accuracy using this formula: $acc_n$ =$l_n$/$m_n \times 100$. This resulted in a list of accuracy scores $ACC$=\{$acc_1$, $acc_2$, ..., $acc_{50}$\}, representing the participants' performance of 50 articles. To further evaluate whether the means of two groups (individual vs. collaborative \& non-experts vs. experts settings) are statistically different, we conducted a paired independent sample T-test. %This article-level test has a sample size equivalent to the total number of articles (n=50). The values for comparison are computed by taking the average performance of participants per article. 
Since the T-test is grounded on the assumption of normality \cite{gerald2018brief}, we ran the Kolmogorov-Smirnov test on our data and confirmed that the requirement was satisfied. Following, we summarize the results of statistical testing.

% \vspace{5pt}
\noindent\textbf{Baseline.}
Each of the 50 3-paragraphed articles has 2 paragraphs authored by human and 1 paragraph deepfake-authored. Therefore, participants have a 1/3 chance of selecting the deepfake
paragraph, and we developed a random generator to randomly identify one of the paragraph as deepfake. As such, the baseline performance of the random-guessing accuracy is approximately to be 33.33\%. 
% \hua{[TODO] Describe Random baseline.}

\begin{table*}
\centering
\begin{adjustbox}{max width=\textwidth}
\begin{tabular}{c|c|c|c|c|c|c|c|c} 
\toprule
\multirow{3}{*}{\textbf{Explanation Type}} & \multicolumn{4}{c|}{\textbf{Correct Detection}} & \multicolumn{4}{c}{\textbf{Incorrect Detection}}  \\
\cmidrule{2-9}
                                    & \multicolumn{2}{c|}{\textbf{Non-Expert (I vs. C)} }  & \multicolumn{2}{c|}{\textbf{Expert (I vs. C)} }                                                 & \multicolumn{2}{c|}{\textbf{Non-Expert (I vs. C)} }   & \multicolumn{2}{c}{\textbf{Expert (I vs. C)} }                                                   \\ 
\cmidrule{2-9}
                                    & \textbf{Percentage (\%) }        & \textbf{P-Value} & \textbf{Percentage (\%)}   & \textbf{P-Value}   & \textbf{Percentage (\%)}   & \textbf{P-Value}  & \textbf{Percentage (\%)}   & \textbf{P-Value}                                                    \\ 
% \cmidrule{1-9}
% \cmidrule{1-9}
\midrule 
\midrule
Grammar     & 13.97 vs. 23.08** & 0.004  & 15.33 vs. 24.6*** & 0.001          & 15.65 vs. 16.89 & 0.587    & 14.22 v.s 12.07 & 0.329                            \\ \cmidrule{1-9}
Repetition   & 6.73 vs. 6.69   & 0.98    & 4 vs. 6.4      & 0.125                                                      & 8.53* vs. 5.62   & 0.044   & 1.67 vs. 2      & 0.703                                                      \\ 
\cmidrule{1-9}
Common Sense                        & 9.25 vs. 15.48*  & 0.011  & 13 vs. 28***      & 4.8e-07       & 13.02 vs. 9.94  & 0.166    & 3.33 vs. 5.56   & 0.163                                                      \\ 
\cmidrule{1-9}
Logical Errors                      & 11.64 vs. 10.24 & 0.496   & 7.78 vs. 14.4**  & 0.002     & 18.54*** vs. 7.7   & 3.3e-05 & 3.89 vs. 4      & 0.938                                                      \\ 
\cmidrule{1-9}
Self-Contradiction                  & 9.35 vs. 5.57   & 0.077   & 7.67 vs. 14.8***  & 0.001        & 18.01*** vs. 6.7   & 9.6e-07 & 6.56 vs. 3.6    & 0.054                                                      \\ 
\cmidrule{1-9}
Lack of Creativity                  & 12.87 vs. 13.49 & 0.776   & 8.33 vs. 7.6   & 0.714      & 16.9 vs. 14.13  & 0.322    & 8.11** v.s 3.6    & 0.002   \\ 
\cmidrule{1-9}
Coherence           & 14.64 vs. 19.29* & 0.045  & 20.56 vs. 32***   & 0.0008                                                    & 11.65 vs. 10.06 & 0.318    & 13.78** vs. 9.2   & 0.003     \\ 
\cmidrule{1-9}
Other                               & 0 vs. 0         & N/A     & 12.22 vs 18.4*  & 0.014                                                     & 0 vs. 0         & N/A      & 6.78 vs. 8.4    & 0.264                                                      \\
\bottomrule
\end{tabular}
\end{adjustbox}
% \captionsetup{justification=centering} % Center the caption
\caption{The percentage of frequency for each reasoning explanation category w.r.t. correct \& incorrect detection (I vs. C = Individual vs. Collaboration) and corresponding t-test results (***: \textit{p} $<$ 0.001, **: \textit{p} $<$ 0.01, *: \textit{p} $<$ 0.05).}
%\vspace{-10pt}
\label{tab:reasoning-t2}
\end{table*}

\subsection{Study 1: Collaboration between AMT Workers}

%\kenneth{Sometimes we said ``Study 1'' but sometimes we said ``Experiment 1''. Let's all just use ``Study 1''-- same applied to Study/Experiment 2.}

%\vspace{5pt}
\paragraph{Detection Performance.}
From Table \ref{tab:rq1_amt} we observe that
English non-experts achieve an average accuracy of
44.99\% individually, which is a 11.66\% increase from the 
baseline (random-guessing) of 33.33\%. 
Using a paired T-test to measure statistical significance, 
the baseline vs. individual performance comparison achieve a p-value
of $3.8e{-}05$ which indicates strong statistical significance. 
Next, for the collaborative setting, the non-experts collaborate 
asynchronously, achieving an average accuracy of 51.35\%.
The p-value of Individual vs. Collaboration comparison is 0.054, 
indicating weak statistical significance. 
However, the comparison of Baseline vs. Collaboration yields a 
p-value of $2.8e{-}05$ which indicates strong significance. 
Thus, all comparison groups for non-experts indicate strong significant improvement, 
except for Individual vs. Collaboration in which the improvement observed during collaboration is weak.

\paragraph{Analysis of Reasoning Explanations.}
In Table \ref{tab:reasoning-t2}, we measure the statistical significance of explanations used by participants individually and collaboratively for each of the seven reasoning explanations, where we divide based on both correct and incorrect detection responses. 
For AMT (\ie non-experts), we observe only a few statistically significant explanations. 
Correct responses show significant scores for grammar, common sense, and coherence.
While incorrect responses have significant scores for 
repetition, logical errors, and self-contradiction. Furthermore, we visualize these explanations for both correct and incorrect 
responses in Figures \ref{fig:rq1_justification_correct} and \ref{fig:rq1_justification_incorrect}, respectively. In these figures, we 
observe that non-experts, both Individually and collaboratively, 
do not show any patterns in response. Thus, in summary, these factors yielded a minimal improvement in performance when non-experts collaborated.  
Another reason for the minimal improvement is the style of collaboration utilized by non-experts - asynchronous collaboration. We further elaborate on the potential hypothesis  in the Discussion section below.
%Section \ref{section:discussion}.

\subsection{Study 2: Collaboration Between Upwork Participants}

%\kenneth{I revise it to ``Study 2'' instead of ``Experiment 2.''}

%\vspace{5pt}
\paragraph{Detection Performance.}
The English experts achieve an average accuracy of 56.11\% 
and a p-value of 8.2e-11 for Baseline vs. Individual, indicating 
strong significance. In the collaborative (synchronous) setting, 
the participants achieve an average accuracy of 68.87\% with 
a p-value of 1.3e-05 for Individual vs. Collaboration, suggesting 
a strong statistical significance. 
Also the p-value for the comparison of Baseline vs. Collaboration 
(1.2e-12) indicates an even stronger significance.

\paragraph{Analysis of Reasoning Explanations.}
In Table \ref{tab:reasoning-t2}, 
we measure the statistical significance of 
explanations used individually and collaboratively. 
We measure two categories when responses are correct 
and incorrect. 
For Upwork (experts), we observe more statistically 
significant explanations for correct responses than 
for incorrect responses. Correct responses 
had 6 statistically significant types from collaborations out of 8 -
grammar, common sense, logical errors, 
self-contradiction, coherence, and other. 
Next, incorrect responses recorded only 2 statistically significant responses - lack of creativity and coherence. Furthermore, we observe in Table \ref{tab:reasoning-t2} that experts show a much stronger 
significance (p-value $<$ 0.01) in the frequency of explanations used than non-experts in the correct detection cases. 

\begin{figure*}[!t]
\centering
\includegraphics[width=0.75\textwidth]{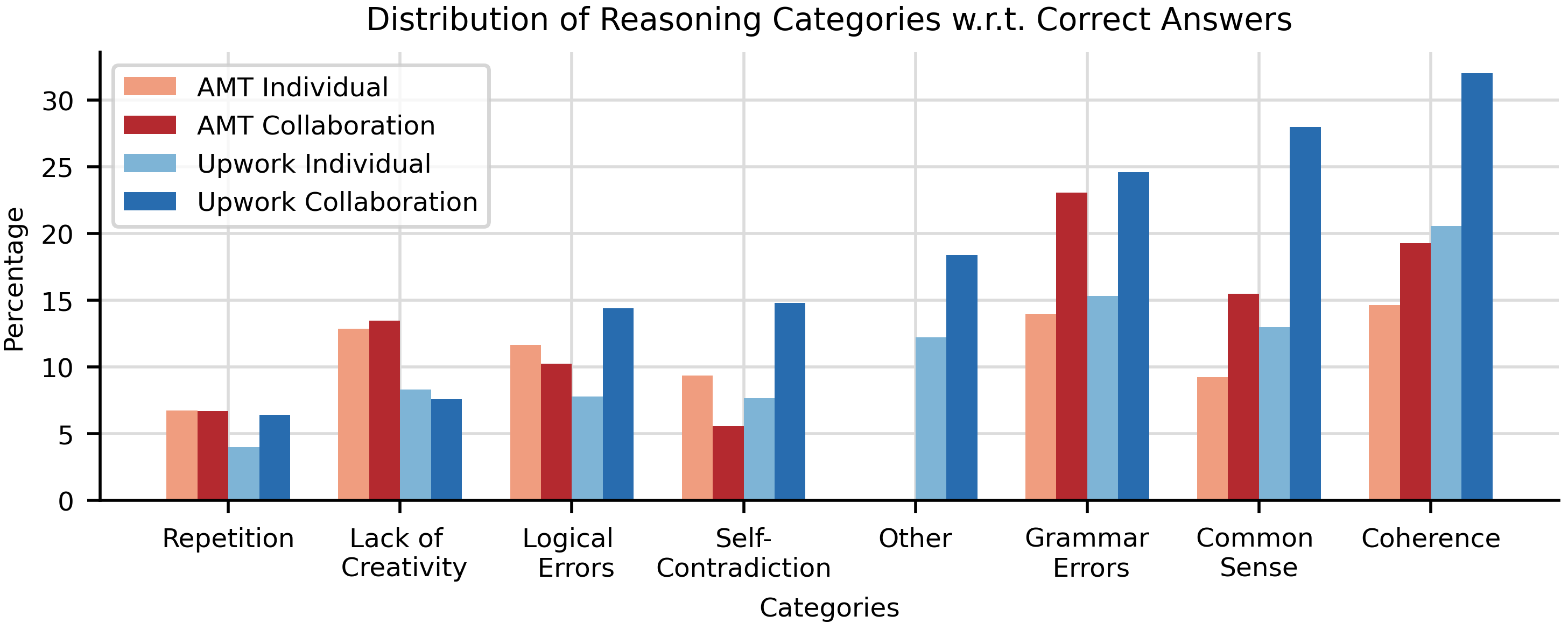}
% \vspace{-8pt}
\caption{The percentage of frequency for selected reasoning explanation w.r.t. correct human detection.}
% \vspace{-8pt}
\label{fig:rq1_justification_correct}
\end{figure*}

\begin{figure*}[h]
\centering
\includegraphics[width=.75\textwidth]{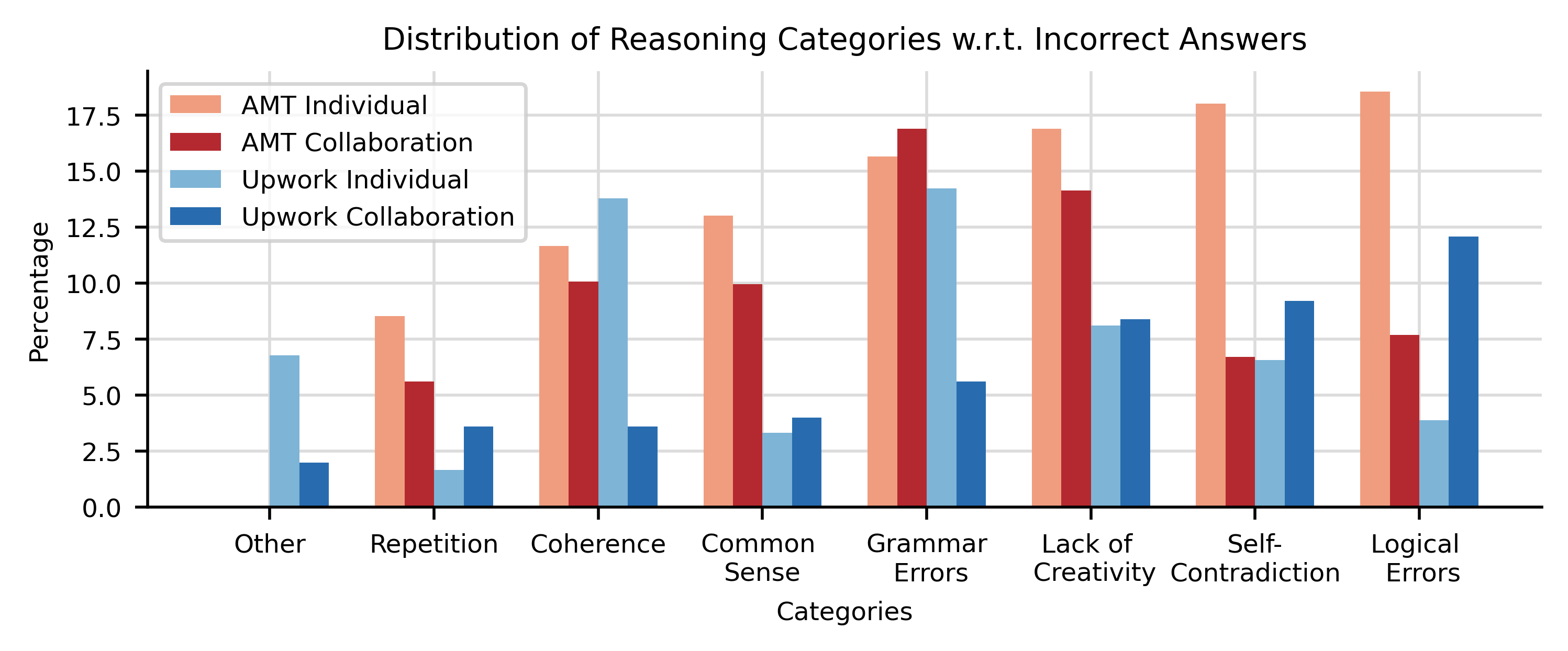}
% \vspace{-10pt}
\caption{The percentage of frequency for selected reasoning explanations w.r.t. incorrect human detection.}
% \vspace{-1em}
\label{fig:rq1_justification_incorrect}
\end{figure*}

Furthermore, Figures \ref{fig:rq1_justification_correct}
and \ref{fig:rq1_justification_incorrect} visualize the 
frequency of explanations used by participants for 
correct and incorrect responses, respectively. 
We observe that experts used coherence, common sense,
grammar errors, other,\footnote{off-topic and off-prompt were the most frequent justifications.} and self-contradiction 
more frequently collaboratively for correct responses.
However, individually, they used grammar errors, coherence 
and other more frequently for incorrect responses. 
% \hua{I complerely agree with coherence and common sense, but I could not understand the self-contradiction well, would you like to teach me more on it?}
This suggests that coherence, common sense, and self-contradiction are strong 
indicators for distinguishing deepfake texts from human-written texts, since they are the only explanations do not overlap in frequency between correct and incorrect responses.

\section{Discussion}
\label{section:discussion}

% \hua{}
% In **Discussion** section, we can potentially discuss three things:
% % 
% 1. hypothesis (potential reasons) of 'why p-value is not stats significant'.
% % 
% 2. Comparing "Non expert vs. Expert", and explicitly say Expert outperforms non-expert, but maybe due to higher payment, synchronous setting, and professional knowledge, but these factors are hard to tease apart. As a result, we suggest the a strategy is to "ask experts to collaboratively detect in a synchronous setting".
% % 
% 3. ChatGPT is even worse than Non-expert, this again, emphasize the task difficulty, and shows that human detection is not replaceable by AI models yet.
% % 
% Please feel free to add more discussion from our previous result lol!
% % 

% \subsection{Implications}

\subsection{Deepfake Text Detection is Non-Trivial for Humans}
% \noindent \textbf{Deepfake text detection is a non-trivial task for humans.}
% \hua{I moved the ChatGPT part into method, as that is really amazing experiments -- I think we should move it forward to emphasize more. Here we can probably roughly summarize all the above results.}
% 
% 
In order to further confirm the difficulty of the task of detecting 1/3 of paragraphs as deepfake, we asked ChatGPT to perform the task. 
Recently, ChatGPT\footnote{https://openai.com/blog/chatgpt}
has been found to have emergent abilities \cite{OpenAI2023GPT4TR}, one of which is being able to 
accurately perform many 
text classification tasks accurately. 
Given the recent observation that using personas with ChatGPT improves accuracy, we use 
a persona-themed prompt - \textit{You are an expert. Which of the 3 paragraphs is AI-generated? Answer choices: paragraph\_1, paragraph\_2, or paragraph\_3}.
Using this prompt, ChatGPT achieves a 38\% accuracy, only 5\% above the baseline. 
In fact, ChatGPT got confused with certain paragraphs that it deviated from 
the answer choices and generated other responses, such as 
none of the paragraphs are AI-generated, 
all are AI-generated, 
or picking two instead of one paragraph, e.g., 
paragraph\_1 \& paragraph\_3. While using our framework, humans achieve an average accuracy of 
44.99\% (non-experts) and 51.35\% (experts), individually. 
Through collaboration, their performances increase 
to an average accuracy of 51.35\% and 68.87\% for non-experts. 

% However, while using our framework, humans achieve an average accuracy of 
% 45\% (non-experts) and 51\% (experts), individually. This observation suggests the non-trivial nature of the task. Through collaboration, their performances increase 
% to an average accuracy of 51\% and 69\% for non-experts. 
% This further reinforces the non-trivial nature of this task. In summary, while the improvement observed 
% through collaboration is 
% laudable, there is still room for improvement, therefore in the future, we need to develop tools such as ones with human-in-the-loop
% to assist humans to perform the task more accurately and effectively.

% As can be seen in the aforementioned results, detecting deepfake texts is a non-trivial task. 
% %Next, we further discuss our hypotheses on the reasons for these observations. 
% Individually, non-expert and expert humans achieve 
% a 45\% and 56\% average accuracy, respectively on the task. 
% Through collaboration, their performances increase 
% to an average accuracy of 51\% and 69\% for non-experts 
% and experts, respectively. Even though the current results outperform baseline with a large margin, they are still far from perfection. 
% Alarmingly, when ChatGPT, an expert NTG is asked to perform the task, it achieves only a 38\% accuracy. 
%\hua{[TODO] Add more contents :)}

\subsection{Detection Performance Comparison Between Experts and Non-Experts}

% \subsection{Collaboration significantly improves performance}
% \subsection{Experts outperforms Non-Experts in Deepfake Detection}

In the aforementioned sections, we observe that collaboration is an effective approach 
% is a statistically significant technique 
to improve human performance in deepfake text detection. 
% This improvement is unsurprising as it has long been established that the performance of a group may surpass that of even the most knowledgeable person \cite{mercier2011humans}. 
% 
% 
Further, the results in Tables \ref{tab:rq1_amt}
and \ref{tab:rq1_Upwork} also suggests that \textbf{experts achieve a more significant improvement with collaboration than non-experts}. There are two potential reasons for this: 
(1) expert participants are able to utilize their expert knowledge more efficiently when collaborating. This is further confirmed in Figures \ref{fig:rq1_justification_correct}, where Upwork (experts) Collaboration 
show more frequent use of coherence, common sense, grammar errors, 
other, self-contradiction, and logical errors than Individually. 
The intuition here is individually, expert participants did not use these explanations as frequently which yielded an average accuracy of 56.11\%, however, during collaboration, they used these explanations more frequently and accurately, improving the average accuracy to 68.87\%.
(2) The second reason is argued by the body of CSCW literature (e.g., \cite{birnholtz2012tracking, shirani1999task, mabrito2006study}) which suggests that the gains of synchronous 
collaboration outweighs the benefits of asynchronous collaboration.

However, for non-experts due to the erratic usage of explanations as observed in Figures \ref{fig:rq1_justification_correct} and \ref{fig:rq1_justification_incorrect},
it is difficult to ascertain a pattern. This is potentially the reason why, 
although collaboration is statistically significant for non-experts, 
it is a weak significance (p-value=0.054). Furthermore, this suggests that while experts are able to collaborate well, non-experts may require a guided 
synchronous collaboration strategy to further improve performance. 
% 
% new added limitation
It is worth noting that when comparing the Non-expert with AMT and Expert with Upwork performance, the difference may potentially also be resulted from different collaboration modes (\ie ``asynchronous'' vs. ``synchronous'' settings) and different compensation levels. However, with the respective rational settings with two groups, the Experts can outperform non-experts in detecting deepfake texts.

\subsection{Which Explanation Categories Can Potentially Be Helpful for Deepfake Text Detection?}

%\hua{Hi Adaku, could you please double check the below contents can answer this question? Thanks!}

% \subsection{Justification patterns are more present with Experts}
% 
% 
Experts' mentions of \textbf{coherence, logical errors, and self-contradiction errors} as explanations for deepfake text detection were significantly higher in the collaborative setting than in the individual setting, specifically for correct responses (Figure \ref{fig:rq1_justification_correct}).
Non-experts showed no pattern differences in coherence, logical errors, and self-contradiction explanations between individuals and collaboration. However, expert participants used them, especially coherence and self-contradiction, more in collaboration when they detected the deepfake texts successfully and less in collaboration when they detected deepfake texts inaccurately. This result corroborates \citet{dou2022gpt}'s finding that machines are prone to fall short of those categories. 
Taking into account experts' superior performance in deepfake text detection, we conclude that both coherence errors and self-contradiction errors are strong indicators of deepfake text. 
Table \ref{tab:reasoning-t2} confirms this finding as well since both explanations 
have a p-value $<$ 0.001 for correct responses, suggesting very strong significance. 
Regarding logical errors, expert participants used them
more frequently in the collaborative setting for both 
correct and incorrect responses. That said, our findings imply that logical flaws may be a weak predictor of deepfake.

\subsection{Which Explanation Categories Should Be Cautious Indicators for Deepfake Text Detection?}
% \textbf{Grammar Errors are not good indicators of deepfake texts.}

%\hua{Hi Adaku, would we have more indicators, more than grammar errors, to add as ``bad indicators'' here?}

In line with previous works~\cite{dou2022gpt,clark-etal-2021-thats}, we observe that participants' mentions of \textbf{grammar errors, repetition, creativity} were associated with incorrect detection of deepfake texts. These markers are identified as deceptive indicators of deepfake texts.

%\kenneth{Prior works have studied this. We should cite them here and state what is new that we found out. Or we just echoed their findings. As this is not our major contribution, echoing prior findings is ok I think.}

\begin{enumerate}
    \item \textbf{Grammar Errors}:
    Experts use grammar errors frequently for both correct 
    (collaboration) and incorrect (individual) responses. 
    %This is because grammar errors cannot be used as 
    %a sole explanation for distinguishing deepfake 
    %texts from human-written texts. 
    This could be attributed to the fact that humans are equally prone to making grammatical mistakes. As a result, employing this explanation can result in both accurate and inaccurate detection. Still, our results indicate that experts can use grammar errors for detection signals more correctly as opposed to non-experts. Non-experts 
    use grammar errors frequently for both incorrect and correct responses, although a bit more frequently for incorrect responses. 
    Furthermore, this phenomenon is 
    confirmed by the findings in \cite{clark-etal-2021-thats, dou2022gpt} that grammar errors are weak indicators of deepfake texts. Therefore, we conclude that grammar errors are weak indicators of deepfake texts.

    \item \textbf{Repetition:}
    We observe in Figure \ref{fig:rq1_justification_correct} and \ref{fig:rq1_justification_incorrect} that repetition 
    is the last and second-last frequently used explanations for correct and incorrect responses, respectively. 
    This was a good indicator of deepfake texts when NTGs were still in their infancy. 
    However, NTGs have improved 
    significantly such that the quality of generations can 
    be misconstrued as human-written. %Even though we use GPT-2, which is currently no longer the SOTA generative language model, 
    In addition, we took measures to ensure high-quality generations, which is discussed in detail in the Method Section. 
    Thus, repetition was not prevalent in the deepfake texts, making repetition a weak indicator of deepfake texts. 

    %This further confirms that repetition is a false indicator of deepfake texts. 

    \item \textbf{Creativity:} 
    News is supposed to be the unbiased reporting of factual events. Therefore, as these events remain non-fiction, news articles are not creative and should not be judged by their level of creativity. 
    This is the reason why experts used creativity very sparingly because English experts they 
    are aware of which style of writing should creative or not-creative. 
    Therefore, experts use repetition as explanation second-to-last for correct responses (Figure \ref{fig:rq1_justification_correct}). 
    Unsurprisingly, experts used creativity a bit more frequently for incorrect responses, with the more frequent usage observed in individuals (Figure \ref{fig:rq1_justification_incorrect}). 
    However, for non-experts, creativity is also used sparingly for correct responses but frequently for incorrect responses. 
    Therefore, due to the frequent usage of creativity for incorrect responses vs. infrequent usage for correct responses, it follows that for the task of detecting deepfake news paragraphs, creativity is a false indicator of deepfake texts. 
\end{enumerate}

% Therefore, for non-experts, we observe more erratic behavior in terms of justification 
% which is why there is no pattern and possibly the reason for the weak 
% statistically significant score. This suggests that while non-experts' performance 
% improves through collaboration, unlike experts, non-experts may require 
% a tool or novel strategy to help them collaborate more efficiently or accurately. 

\section{Limitation}
\label{section:limitation}

% \hua{Seem we don't describe future work here? so I just removed ``Future Work'' in the section title? -- it's fine to have no.}
% Why did we only use GPT-2
% Compare AMT non-experts vs AMT experts
% Compare Upwork non-experts vs. Upwork experts 
% Bias in Write setting (framework is not conducive to this scenario)
% 

% The ultimate goal of this research study is two-fold, to answer 
% the questions: (1) does collaboration improve human performance in 
% deepfake text detection? (2) are English experts better at deepfake text detection than English non-experts? These questions require a 
% carefully crafted experimental framework for investigation. 
% This is both financially expensive and time-consuming. 
\noindent
To implement design choices and run manageable experiments, we made a few simplifications that may limit our findings. 
First, since we only use GPT-2
%To mitigate these issues, while preserving the integrity of this research study, we only use one of the best NTGs - GPT-2 
to generate  deepfake texts, our findings may not be directly applicable to other NTGs.
However, our choice of using GPT-2 is reasonable because: (1) prior research reported that human detection performance of deepfake texts by  the later GPT-3 and GPT-2 is similar    \cite{uchendu2021turingbench, clark-etal-2021-thats}, and (2) using the largest parameter size of GPT-2 enabled us to generate deepfake texts more effectively that closely resembles GPT-3 quality. 
Furthermore, as we use the default hyperparameters 
of GPT-2 to generate the texts, the results may be limited to that sampling technique. 
However, we mitigated this issue by 
manually checking the quality of a few of the articles
and found the deepfake texts to be coherent and consistent 
with the rest of the paragraphs. 
%\thai{can we say ``...deepfake texts to be in high quality" instead?}. 
This preserved the integrity of the experiments as the 
task remained non-trivial. 

\section{Conclusion}
\label{section:conclusion}
% We studied human performance in deepfake text 
% detection. 
% To be more realistic, we built a 3-paragraph article with 1/3 paragraphs, LLM-generated (deepfake) and 
% 2/3 paragraphs, human-written. 
% We ask human participants to select which paragraph is deepfake and to provide explanation for their selection 
% out of 7 error types. 
% Specifically, we studied human performance can be improved by collaboration. To achieve this, 
% we recruit non-expert human participants from AMT and experts from Upwork. Furthermore, we run asynchronous collaboration with AMT and synchronous collaboration with Upwork. 
% Finally, our results suggest that collaboration significantly improves performance 
% for both non-experts and experts.  
% We further identify several strong and weak indicators of deepfake texts based on 
% the analysis of explanations used in both correct and incorrect responses. 
% Finally, the enhanced performance of participants from baseline in the individual setting indicates that our 
% \textit{Turing Test} framework facilitates the improvement of humans' deepfake text detection performance. 

Our study investigated the impact of human collaboration on improving the detection of deepfake texts. To create a realistic experimental setup, we constructed a three-paragraph article comprising one LLM-generated (deepfake) paragraph and two human-written paragraphs. Participants were tasked with identifying the deepfake paragraph and providing explanations based on seven explanation types. 
For participant recruitment, we recruited non-expert participants from AMT for asynchronous collaboration and experts from Upwork for synchronous collaboration. 
The results revealed that collaboration is likely to enhance the detection performance of both non-experts and experts. We further identified several strong and weak indicators of deepfake texts through the explanation analysis. Notably, the improved performance of participants compared to the baselines indicated that our \textit{Turing Test} framework effectively facilitated the enhancement of human deepfake text detection performance.

\section{Ethical Statement}
\label{section:ethics}

% \subsection{Ethical Statement}
Our research protocol was approved by the Institutional
Review Board (IRB) at our institution.
%The Pennsylvania State University.
We only recruited human participants 18 years old or over. 
Participants did not have to complete the entire task to be paid. %, as partial completion was acceptable. 
Using AMT, participants' identification was already anonymized, but 
for Upwork we anonymized participants by assigning them numerical values for the analysis. 
For performing the deepfake text detection task, all our human participants, from both AMT and Upwork, were  paid  over minimum wage rate. 
Next, the articles that we used for the experiments are the first 3-paragraphs of news articles. While we did not share the answer to the task, we clearly informed participants that the presented texts (and one of three paragraphs therein) contains deepfake texts.
%so I don’t think that there is a risk that it will be effective fake news. Also, as the goal of the deepfake texts is to be consistent with the rest of the article, anything that passes for fake news will be incoherent. 
Therefore, we believe that participants are unlikely to be negatively influenced by their exposure to the test news articles with deepfake paragraphs.

%could not use our articles maliciously to spread fake news. 

\section{Acknowledgments}
\label{section:ack}
This work was in part supported by NSF awards \#1820609, \#2114824, and \#2131144, and PSU CSRE seed grant 2023.
We thank the crowd workers from AMT and the experts from Upwork for participating in this study.
We also thank the anonymous reviewers for their constructive feedback.

\bibliography{bibliography}

\end{document}